\documentclass[nohyperref]{article}

\usepackage{microtype}
\usepackage{graphicx}
\usepackage{subfigure}
\usepackage{booktabs} % for professional tables

\usepackage{hyperref}
\usepackage{xcolor}

\usepackage[accepted]{icml2022}

\usepackage{amsmath}
\usepackage{amssymb}
\usepackage{mathtools}
\usepackage{amsthm}

\usepackage[capitalize,noabbrev]{cleveref}

\theoremstyle{plain}

\theoremstyle{definition}

\theoremstyle{remark}

\usepackage[textsize=tiny]{todonotes}

\icmltitlerunning{Robustar: Interactive Toolbox for Robust Vision Classification}

\begin{document}

\twocolumn[
\icmltitle{\includegraphics[width=0.15\textwidth]{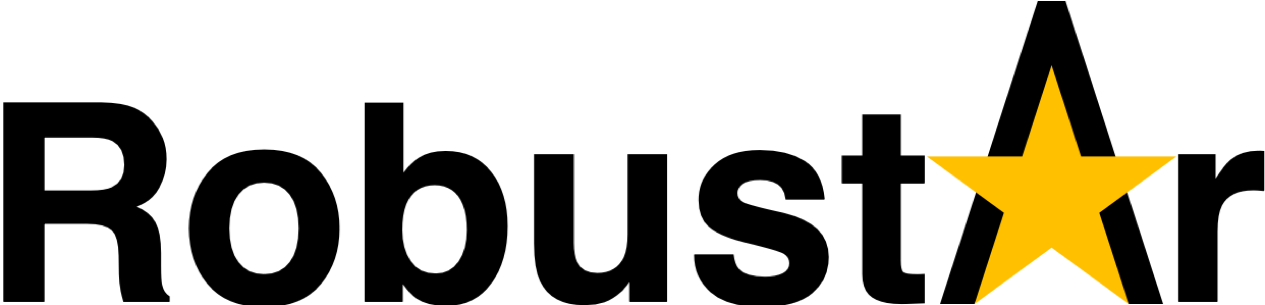}: Interactive Toolbox Supporting Precise Data Annotation \\for Robust Vision Learning}



\icmlsetsymbol{equal}{*}

\begin{icmlauthorlist}
\icmlauthor{Chonghan Chen}{equal,cmu}
\icmlauthor{Haohan Wang}{equal,cmu,uiuc}
\icmlauthor{Leyang Hu}{noc}
\icmlauthor{Yuhao Zhang}{nott}
\icmlauthor{Shuguang Lyu}{uci}
\icmlauthor{Jingcheng Wu}{cmu}
\icmlauthor{Xinnuo Li}{umich}
\icmlauthor{Linjing Sun}{noc}
\icmlauthor{Eric P. Xing}{cmu,mbzuai}
\end{icmlauthorlist}

\icmlaffiliation{cmu}{School of Computer Science, Carnegie Mellon University}
\icmlaffiliation{uiuc}{School of Information Science, University of Illinois Urbana-Champaign}
\icmlaffiliation{noc}{School of Computer Science, University of Nottingham Ningbo China}
\icmlaffiliation{uci}{Bren School of Information and Computer Sciences University of California, Irvine}
\icmlaffiliation{umich}{College of Literature, Science and the Arts, University of Michigan}
\icmlaffiliation{nott}{School of Computer Science, University of Nottingham}
\icmlaffiliation{mbzuai}{Mohamed bin Zayed University of Artificial Intelligence}

\icmlcorrespondingauthor{Robustar Github Repository}{\href{https://github.com/HaohanWang/Robustar}{https://github.com/HaohanWang/Robustar}}

\icmlkeywords{Machine Learning, ICML}

\vskip 0.3in
]



\printAffiliationsAndNotice{\icmlEqualContribution} % otherwise use the standard text.

\begin{abstract}
We introduce the initial release of our software Robustar,
which aims to improve the robustness of vision classification machine learning models through 
a data-driven perspective. 
Building upon the recent understanding that 
the lack of machine learning model's robustness 
is the tendency of the model's learning 
of spurious features, 
we aim to solve this problem 
from its root at the data perspective
by removing the spurious features 
from the data before training. 
In particular, we 
introduce a software that helps the users to 
better prepare the data for 
training image classification models 
by allowing the users to 
annotate the spurious features 
at the pixel level of images. 
To facilitate this process, 
our software also leverages recent advances 
to help identify potential 
images and pixels worthy of attention
and to continue the training with 
newly annotated data. 
Our software is hosted at the 
GitHub Repository 
\href{https://github.com/HaohanWang/Robustar}{https://github.com/HaohanWang/Robustar}. 
\end{abstract}

\section{Introduction}
Machine learning has achieved 
remarkable performances over \textit{i.i.d} benchmarks 
in recent years, 
which has greatly encouraged the community to 
test the methods 
in other scenarios beyond the \textit{i.i.d} settings, 
leading to multiple other 
topics such as 
the study of cross-domain robustness 
including 
domain adaptation \citep{ben2007analysis,ben2010theory}, 
domain generalization \citep{muandet2013domain}, 
and the recent advances beyond the settings 
\citep{ye2021ood,huang2022two}, 
as well as the study of 
robustness against predefined perturbations 
such as adversarial robustness
\citep{SzegedyZSBEGF13,GoodfellowSS14}. 

While there is a diverse set of topics regarding 
machine learning robustness over different directions, 
recently literature suggests 
that a central theme of these 
challenges is the existence of spurious features 
that are not semantics but associated with the labels
\citep{wang2020high}.  
Correspondingly, a proliferation 
of machine learning robustness methods
are designed 
by explicitly or implicitly 
countering the model's tendency 
in learning spurious features \citep{wang2021toward}. 

While the central theme 
of countering the learning of spurious features 
can potentially guide the development of 
many robust machine learning methods \citep{wang2021toward}, 
we nonetheless consider the 
repeated 
case-by-case design and implementation 
of machine learning methods 
over different applications 
might be inconvenient at certain scenarios. 

Therefore, in this project, 
we aim to solve this problem 
in the scope of image classification 
with a model-free manner, 
by focusing on the data perspective 
at the image pixel level. 
We introduce a software that 
allow the users or domain experts 
to annotate the pixels that 
are associated with the 
label in the spurious manner. 
The annotation will allow 
the system to continue to train the model 
with established robust learning methods 
of data augmentation strategies. 

In summary, we introduce the software Robustar
with the following functions:
\begin{itemize}
    \item Core Function: to allow users interact with the training samples, annotating features that are useless used potentially used by the models to help train a robust model to the request of the domain expert. 
    \item Supporting Functions:
    \begin{itemize}
        \item We use influence function to help identify the samples that need attention. 
        \item We use saliency-map style interpretation to help the user identify the features need attention. 
        \item We use segmentation models to help the user select certain pixels more efficiently. 
        \item We use data augmentation and regularizations to help train models invariant to the identified pixels.
    \end{itemize}
\end{itemize}

The remainder of this manuscript will be structured 
as follows. 
We will first offer an overview of our system 
and the main pipeline of the use case
in Section~\ref{sec:overview}. 
Then in Section~\ref{sec:func}, 
we will introduce 
the core functions 
presented in our software. 
Finally, 
we will offer some discussions 
before we conclude in Section~\ref{sec:con}.

\section{System Overview}
\label{sec:overview}
\begin{figure}[t]
    \centering
    \includegraphics[width=.48\textwidth]{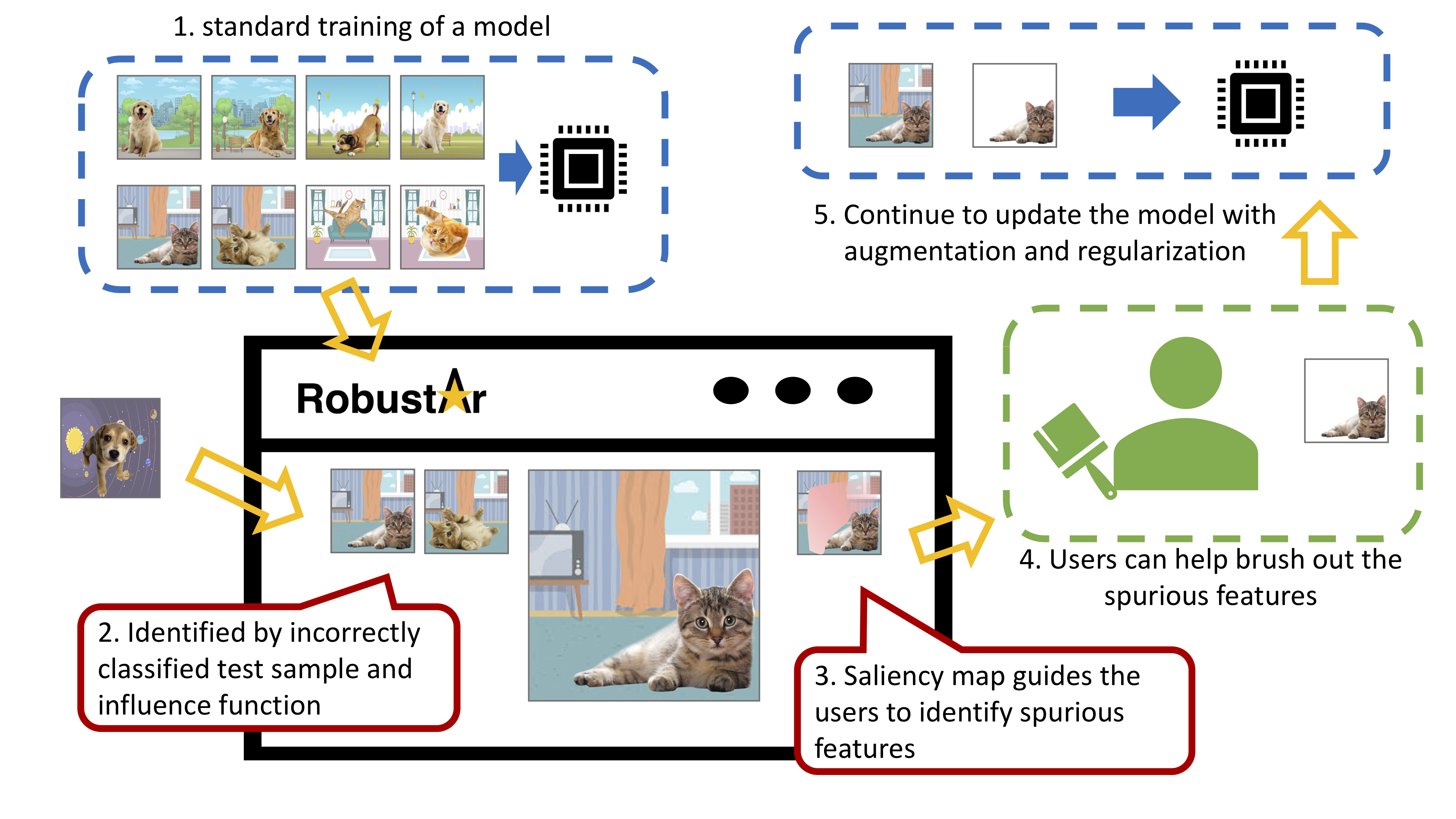}
    \caption{The major working flow of Robustar.}
    \label{fig:robustar:overview}
\end{figure}

Figure~\ref{fig:robustar:overview} shows the overview of the software Robustar pipeline, 
which consists of five major steps. 

\begin{enumerate}
    \item The model can be trained elsewhere and then fed into the software. 
    \item With new test samples, the model can help identify the samples that are responsible for the prediction through influence function. 
    \item The software offers saliency map to help the user know which part of the features the model are paying necessary attention. 
    \item The users can use the drawing tools to brush out the spurious pixels. 
    \item New annotation of these images will serve as the role as augmented images for continued training.
\end{enumerate}

The software can work with models trained elsewhere, as long as these are standard vision models and have the standard pytorch checkpoints available. 
If the influence function results are also calculated, the software can guide the users to the samples that are believed to be responsible for the incorrect classification of the images. Further, our software will also guide the user's attention of the features by the saliency-style interpretation of the models. Then the users can user canvas tool to annotate the spurious pixels of the images. 
Finally, with the newly annotated spurious features, we can continue 
to update the model with data augmentation and consistency regularization to help us discard these spurious features. 

\section{Major Functions}
\label{sec:func}
In this section, we will introduce the 
functions our software offer 
following the natural order of the pipeline 
of our package will 
help the domain experts to annotate the 
spurious pixels in image classification. 

\begin{figure*}[t]
    \centering
    \includegraphics[width=0.9\textwidth]{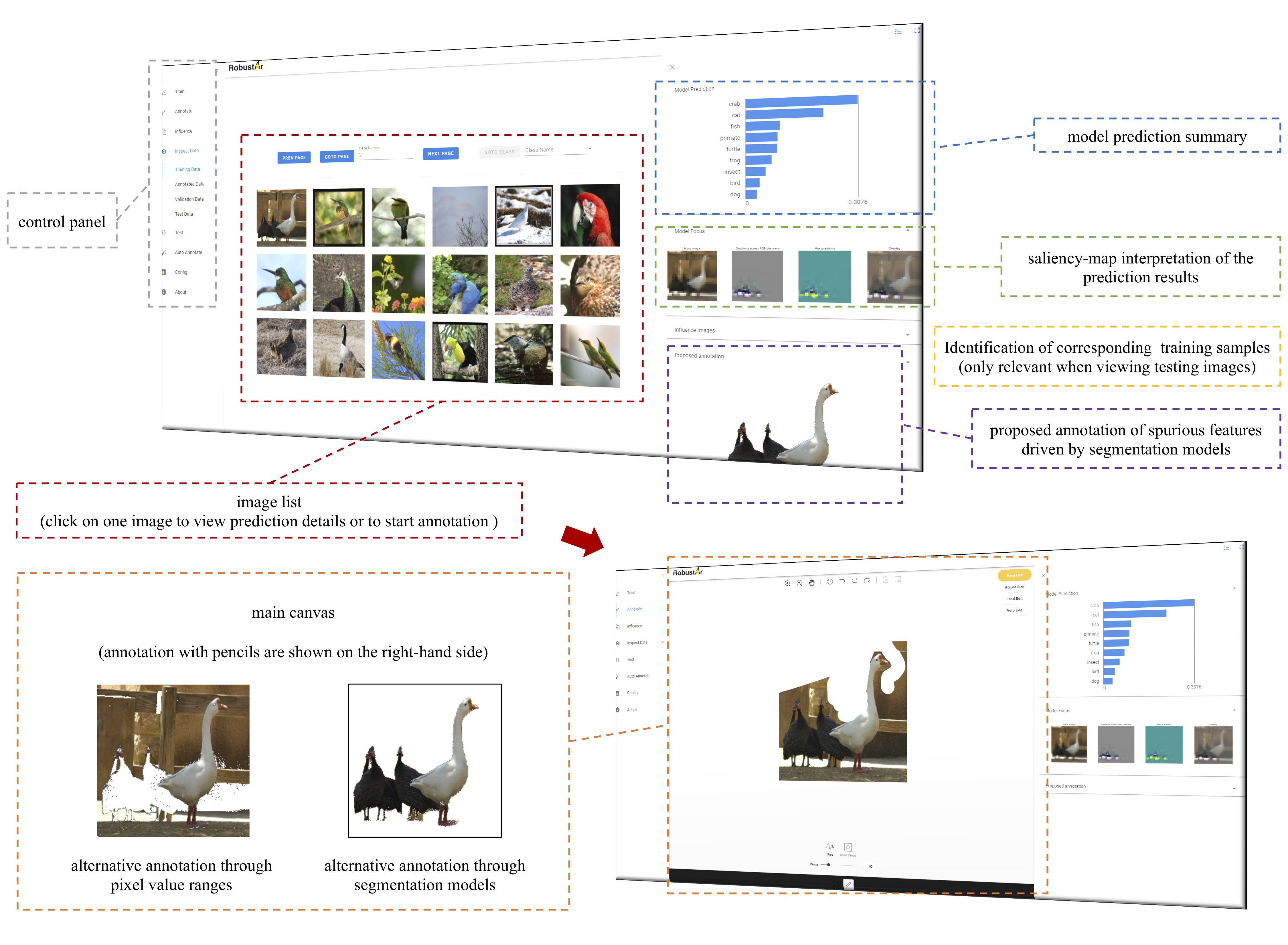}
    \caption{Screenshots of main functions of Robustar: above: the image list to view training images and their role in the model; below: the canvas to annotate the details of one image.}
    \label{fig:screenshot}
\end{figure*}

\paragraph{Image Browsing}
Our system allows the users 
to examine every samples in the training set through 
the image browsing interface (Figure~\ref{fig:screenshot},(upper)). 

Through the interface, the users
can directly view the each image from either the training 
or the testing set. 
If the user click one image, 
the sidebar on the right-hand-side will display multiple 
relevant statistics of the image of interest. 
For example, 
the prediction summary will indicate 
what the model believes the current image to be, 
offering the users the information of 
whether the prediction align with the user's p
perception of the images well. 
The remaining statistics at the sidebar will be discussed later. 

This interface allows the user to check every image in the training sample and identify the ones with spurious features for further annotation.

\paragraph{Automatic Identifying Misleading Samples}
Ideally, to achieve a robust learning system, 
the domain experts will scrutinize 
the training samples 
annotate any spurious pixel features. 
However, 
this examination process 
might potentially 
require an unrealistic amount of working load, 
despite it 
potentially significantly amount of robustness gain. 
Thus, 
to suit the need for some users' 
concerns in devoting the efforts of examining every samples,
we allow the system to propose the suspiciously misleading samples (samples with spurious features learned by the model) first. 

The central assumption that enables the automatic proposal
of misleading samples 
is that 
the spurious features are not shared between training samples 
and testing samples, 
so that when a test sample is misclassified, 
it is mostly due to the fact that the model learns a biased signal that accounts for the 
misclassification. 
This allows us to identify the samples with spurious features.
Influence function \citep{koh2017understanding} conveniently allows us to identify the samples that account for the misclassification:
for every test samples, 
we calculate the most relevant training samples 
from the training set, 
and these samples are proposed to the users 
to pay particular attention 
especially when the test samples 
are misclassified.

\paragraph{Annotation of Spurious Pixel Features}

The user can choose to enter the annotation page
(Figure~\ref{fig:screenshot} (bottom right))
by clicking on any images of interest. 
In the annotation page, the user can 
choose to use pencil to brush out the 
pixels that are considered spurious. 

In addition, to reduce the efforts for user
to identify and brushing out, 
the system offers two other alternative 
ways of annotating the images, 
as shown in (Figure~\ref{fig:screenshot} (bottom left)):
the users can choose to 
directly filter out all the pixels 
with certain ranges, 
we also offer a
pretrained segmentation 
model \citep{he2017mask}
that can 
automatically filter out the background pixels. 
In our experiments, 
we notice that the segmentation 
model can save significant efforts 
in natural images, although not perfect. 
However, we also notice that, in medical images such as X-ray, 
the segmentation model is not always able to reduce 
a significant amount of efforts.

\paragraph{Interpret Model's Decision}

While it will be better to ask the user 
to identify all the pixels in the background, 
it might not be required for 
all the background pixels to be annotated for continued training: 
only the pixels that are learned by the machine learning 
model 
as spurious features need to be annotated
and later countered through the continued training process. 
Therefore, 
to further reduce the users' efforts in annotation, 
the system to show 
where the current model focuses on the images (Figure~\ref{fig:screenshot}). 

% With the identification of misleading samples, 
% we continue to understand which part of the image sample deceives the model's decision. 
% This process requires the neural network interpretation methods. 
% We notice that, if the model is invariant to simple texture variations (\textit{e.g.}, boosted by adversarial training), 
% the pioneering model-interpretation methods, 
% activation maximization \citep{erhan2009visualizing} can sufficiently fulfill our goal. 
For the visualization of the model's decision, 
we simply uses the pioneering model-interpretation methods, 
activation maximization \citep{erhan2009visualizing}, 
and we notice that the method 
can fulfill our need well enough, 
especially when the model is already trained 
to be invariant to high-frequency signals of the images.

\begin{figure}[t]
    \centering
    \includegraphics[width=0.45\textwidth]{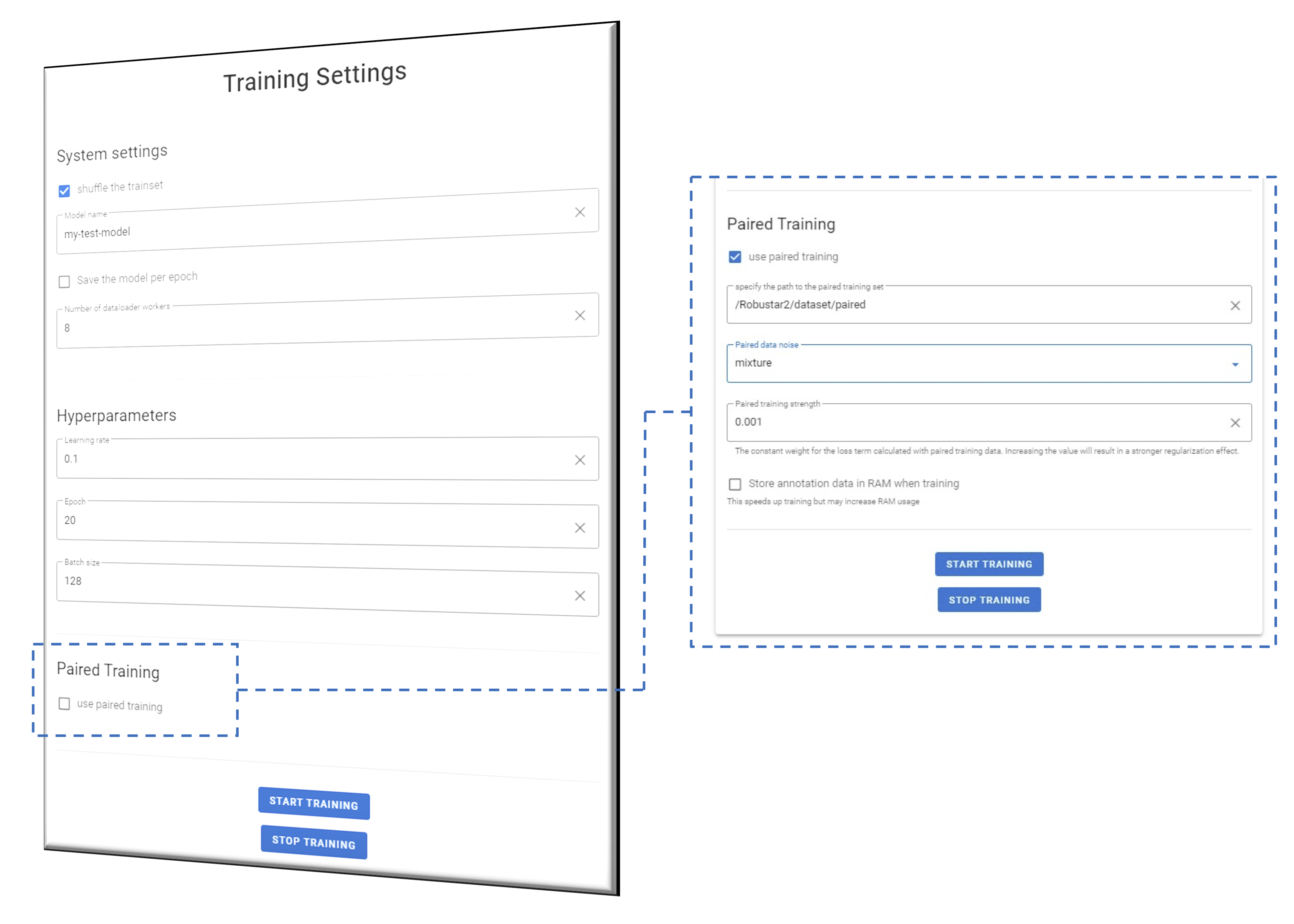}
    \caption{Task center and paired training}
    \label{fig:task}
\end{figure}

\paragraph{Continued Training with Randomized Pixels}
With annotation of spurious features
all misleading samples,
the users can directly use 
our system to update the model 
to improve its robustness against its tendency 
in learning the spurious features
through our task center (Figure~\ref{fig:task}). 

In particular, 
to sufficiently force the model to drop the learning 
of spurious features, 
we leverage the recent training paradigm 
with data augmentation and consistency regularization 
\citep{wang2022toward}:
to train the model with the original image
and an augmented image whose spurious pixel features
are replaced by random noises
and a regularization forcing the model's 
logits (pre-softmax embedding)
to be the same from both set of images.

We refer to this training paradigm 
as paired training. 
\citet{wang2022toward} showed that 
this generic training paradigm can efficiently improve the 
robustness of models empirically better 
than models specifically designed for the tested applications, 
with significantly less computing costs.  

This software paper omits the demonstration 
of the machine learning performances 
through these techniques, 
but one can refer to 
\citep{wang2022toward}
for the superior performances 
led by data augmentation 
and alignment regularization. 

\section{Usage Instructions}
\label{sec:usage}
\paragraph{Preparation}
As the goal of this system is to counter the model's tendency in learning spurious features,
we recommend the users to start with a model with reasonably small training and testing errors. 
Users can either upload such a model to the system 
or use the functions offered by the system to start from scratch and train a new model. 
In either case, the whole training dataset
is expected to be available for the annotation. 

To take the most advantage of the system, we 
recommend the user to use out-of-domain test samples 
(\textit{i.e.}, test samples from an independent data collection instead of from a cross-validated split), 
because the test data from the same collection (distribution) with training data tend to share the same spurious features with training data, 
thus may be inadequate to help identify the spurious features. 
As long as the test samples can represent a wider range of distributions, we do not need a large number of test samples. 

\paragraph{Setup and Run}
With \href{https://www.docker.com/}{Docker} installed, the users can run Robustar 
with a single-line command, 
if the user has pulled the relevant docker images 
with another command line, 
which is only required before the first run. 

To run the system, 
the users need to feed in several path 
for the Docker to mount, 
including the folders of the data 
(with \href{https://pytorch.org/vision/stable/generated/torchvision.datasets.ImageFolder.html}{PyTorch ImageFolder} structure), 
folder of influence images (precomputed or empty folder path), 
folder of checkpoints, 
and a configuration file
with information such as number of classes 
and model configuration. 
Details of these information can be found at the  \href{https://github.com/HaohanWang/Robustar}{Robustar GitHub Repository}. 

\begin{figure}
    \centering
    \includegraphics[width=0.45\textwidth]{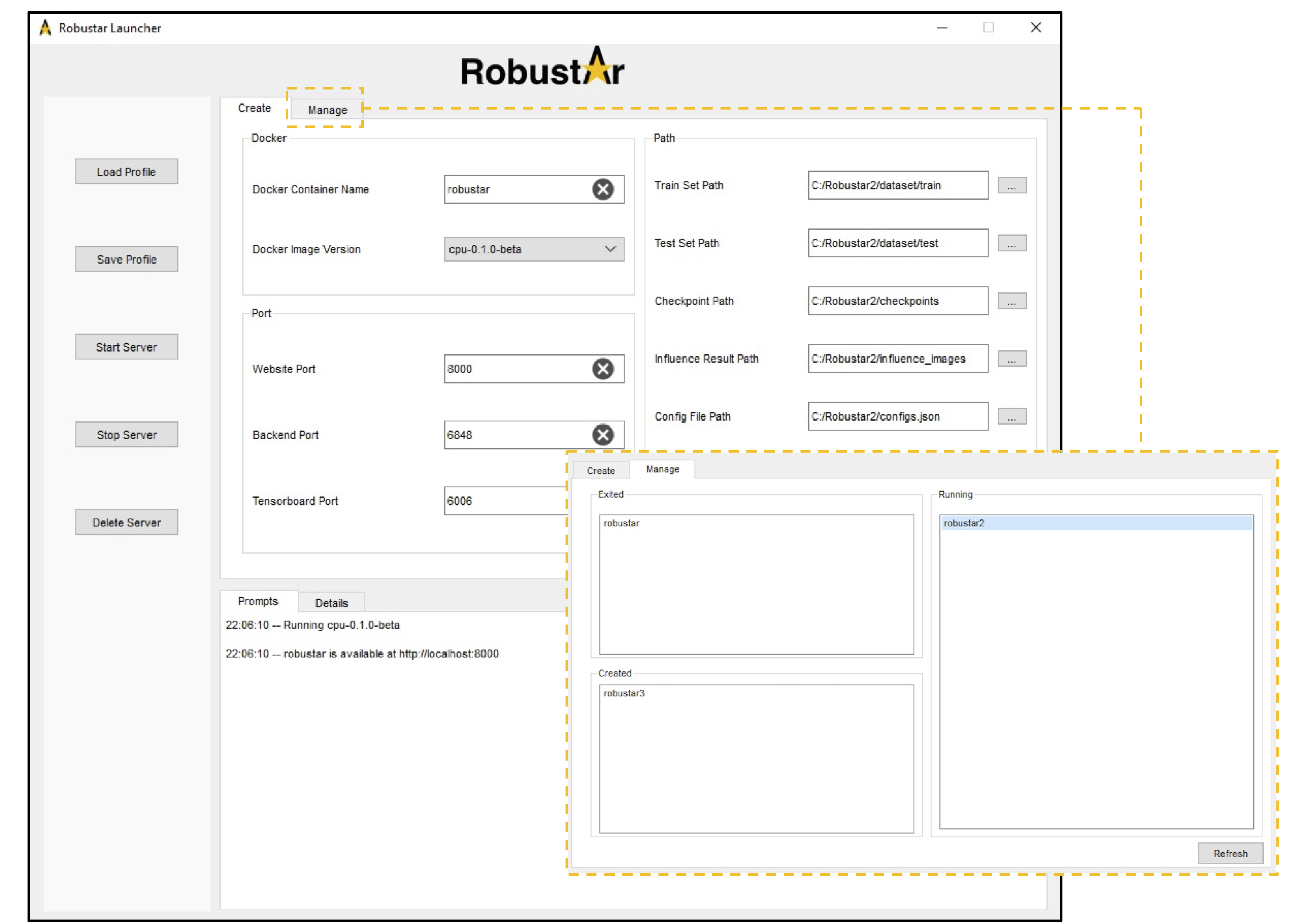}
    \caption{GUI console}
    \label{fig:launcher}
\end{figure}

\paragraph{GUI Console}
For users not comfortable using command line, 
we also offer a GUI console 
for the users to setup the above the configurations
and monitor the running of the system (Figure~\ref{fig:launcher}). 

% \section{Post-Annotation}
% After the annotation, 
% the system will continue to update the 
% model through data augmentation 
% and alignment regularization 
% (consistency loss)
% with the AlignReg package
% introduced in \citep{wang2022toward}

% \section{Discussion}
% \label{sec:discussion}
% \input{secs/discuss}

\section{Discussion and Conclusion}
\label{sec:con}
% Machine learning models' tendency in learning 
% spurious feature is one of the major 
% reasons accounting for model's lack of robustness. 
% In this project, 
We aim to improve machine learning robustness from its root at the data perspective by allowing the users 
to train a model with spurious features annotated. 
For this purpose,
we offer a software for the users to inspect the training data and annotate the spurious image features. 
% Further, our software can also help to identify the samples that may mislead the model and can help pinpoint the features that are used by the model to reduce the workload of the users. 
Our software can be found as \href{https://github.com/HaohanWang/Robustar}{https://github.com/HaohanWang/Robustar}. 
% With Docker installed, our software can be installed and run with one command. 

In addition to the facilitate machine learning robustness, 
we believe our system can 
also serve multiple purposes
in the machine learning community 
from the data perspective. 
For example, 
we expect our system to also help the 
community to inspect 
the properties of the data 
to understand how to annotate and prepare the data
better at the pixel level, 
and our software has the potential 
to generate massive amount of data 
along the interactive process between 
human and the model, 
leading to another possibility 
to automatically understand the properties of the data.

\section*{Acknowledgements}
The project team would like to thank Donglin Chen for his contribution at the early-stage of the development.

\bibliography{ref}
\bibliographystyle{icml2022}

%%%%%%%%%%%%%%%%%%%%%%%%%%%%%%%%%%%%%%%%%%%%%%%%%%%%%%%%%%%%%%%%%%%%%%%%%%%%%%%
%%%%%%%%%%%%%%%%%%%%%%%%%%%%%%%%%%%%%%%%%%%%%%%%%%%%%%%%%%%%%%%%%%%%%%%%%%%%%%%
% APPENDIX
%%%%%%%%%%%%%%%%%%%%%%%%%%%%%%%%%%%%%%%%%%%%%%%%%%%%%%%%%%%%%%%%%%%%%%%%%%%%%%%
%%%%%%%%%%%%%%%%%%%%%%%%%%%%%%%%%%%%%%%%%%%%%%%%%%%%%%%%%%%%%%%%%%%%%%%%%%%%%%%
% \newpage
% \appendix
% \onecolumn

% You can have as much text here as you want. The main body must be at most $8$ pages long.
% For the final version, one more page can be added.
% If you want, you can use an appendix like this one, even using the one-column format.
%%%%%%%%%%%%%%%%%%%%%%%%%%%%%%%%%%%%%%%%%%%%%%%%%%%%%%%%%%%%%%%%%%%%%%%%%%%%%%%
%%%%%%%%%%%%%%%%%%%%%%%%%%%%%%%%%%%%%%%%%%%%%%%%%%%%%%%%%%%%%%%%%%%%%%%%%%%%%%%

\end{document}